\documentclass[aps,prx,twocolumn,superscriptaddress,a4paper,english]{revtex4-1}

\usepackage{times}
\usepackage{color}
\usepackage[colorlinks=true,urlcolor=blue,citecolor=blue]{hyperref}
\usepackage{url}
\usepackage{breakurl}
\usepackage{soul}
\usepackage{graphicx}
\usepackage{amsmath}
\usepackage{subfigure}
\usepackage{xcolor}
\usepackage{amssymb}
\usepackage{latexsym}
\usepackage{hyperref}
\DeclareGraphicsExtensions{.png,.eps}
\usepackage{float}

\usepackage{xcolor}
\usepackage[urlcolor=blue]{hyperref}      
\hypersetup{
    colorlinks = true,                    
    citecolor = {blue},
    linkcolor = {purple},
           }
\begin{document}	
	
\title{Quantum Compressed Sensing with Unsupervised Tensor-Network Machine Learning}

\author{Shi-Ju Ran}\email[Corresponding author. Email: ] {sjran@cnu.edu.cn}
\affiliation{Department of Physics, Capital Normal University, Beijing 100048, China}
\author{Zheng-Zhi Sun}
\affiliation{School of Physical Sciences, University of Chinese Academy of Sciences, P. O. Box 4588, Beijing 100049, China}
\author{Shao-Ming Fei}
\affiliation{School of Mathematical Sciences, Capital Normal University, Beijing 100048, China}
\affiliation{Max-Planck-Institute for Mathematics in the Sciences, 04103, Leipzig, Germany}
\author{Gang Su}
\affiliation{School of Physical Sciences, University of Chinese Academy of Sciences, P. O. Box 4588, Beijing 100049, China}
\affiliation{Kavli Institute for Theoretical Sciences, and CAS Center for Excellence in Topological Quantum Computation, University of Chinese Academy of Sciences, Beijing 100190, China}
\author{Maciej Lewenstein}
\affiliation{ICFO-Institut de Ciencies Fotoniques, The Barcelona Institute of Science and Technology, 08860 Castelldefels (Barcelona), Spain}
\affiliation{ICREA, Passeig Llu\'is Companys 23, 08010 Barcelona, Spain}
\date{\today}

\begin{abstract}
	We propose tensor-network compressed sensing (TNCS) by combining the ideas of compressed sensing, tensor network (TN), and machine learning, which permits novel and efficient quantum communications of realistic data. The strategy is to use the unsupervised TN machine learning algorithm to obtain the entangled state $|\Psi \rangle$ that describes the probability distribution of a huge amount of classical information considered to be communicated. To transfer a specific piece of information with $|\Psi \rangle$, our proposal is to encode such information in the separable state with the minimal distance to the measured state $|\Phi \rangle$ that is obtained by partially measuring on $|\Psi \rangle$ in a designed way. To this end, a measuring protocol analogous to the compressed sensing with neural-network machine learning is suggested, where the measurements are designed to minimize uncertainty of information from the probability distribution given by $|\Phi \rangle$. In this way, those who have $|\Phi \rangle$ can reliably access the information by simply measuring on $|\Phi \rangle$. We propose q-sparsity to characterize the sparsity of quantum states and the efficiency of the quantum communications by TNCS. The high q-sparsity is essentially due to the fact that the TN states describing nicely the probability distribution obey the area law of entanglement entropy. Testing on realistic datasets (hand-written digits and fashion images), TNCS is shown to possess high efficiency and accuracy, where the security of communications is guaranteed by the fundamental quantum principles.
\end{abstract}

\maketitle

\section{Introduction}

An important perspective of quantum information is to transfer and process classical information by taking advantage of quantum physics. Taking dense/super-dense coding protocol \cite{BW92DenseCode, LLTL02densecode, LAH02densecode, BDL04densecode, HHL04densecode, WDLLL05densecode, PPA05densecode} as an example, the idea is to use previously shared entangled state between a sender and the receiver(s) to send more classical information than is possible without the resource of entanglement.  Another example is the machine learning by tensor network (TN) \cite{SS16TNML, LRWP+17MLTN, HWFWZ17MPSML, GJLP18MPOML, GPC18gTNML, S18MERAML, HPWS18TNQML, CWXZ19generateTTNML}. The aim is to employ TN (see some reviews of TN in Refs. \cite{VMC08MPSPEPSRev, CV09TNSRev, S11DMRGRev, O14TNSRev, HV17TMTNrev, RTPC+17TNrev}) as a novel machine-learning model to learn, classify, and/or generate classical information in the quantum many-body Hilbert space. Similarly, classical techniques can assist quantum approaches. One example is to use compressed sensing \cite{D06CS} (see also the book in Ref. \cite{EK12CSbook}) to improve quantum state tomography \cite{GLFBE10STCS, FGLE12QSCS, SRAS+13QSCS, KKD15QSCS}.

We here combine the ideas of compressed sensing \cite{D06CS}, quantum communication \cite{NC02Qcmp}, and unsupervised TN machine learning \cite{HWFWZ17MPSML}, aiming at developing novel quantum communication schemes. Compressed sensing is a powerful scheme for classical data compression by sampling, which is particularly useful when the samplings of the data are difficult or expensive. For instance in the magnetic resonance imaging, compressed sensing can largely compress the required samplings, thus significantly improve the efficiency \cite{lustig2008compressed}. In quantum communication, measurements are also expensive, since quantum states are difficult to prepare and each measurement will collapse or disturb the state. Consequently, the quantum communications of realistic data (e.g., images of $O(10^2)$ bits or more), even including the corresponding simulations of the quantum processes on classical computers, are extremely challenging. Recently, booming progresses have been made in TN machine learning, with which the realistic data (e.g., hand-written digits and photos of articles) can be processed and analyzed by quantum approaches (see \cite{SS16TNML} for instance). High efficiencies haven been demonstrated at least for the classical simulations of these quantum processes. These achievements allow and motivate to develop novel quantum schemes that could not be efficiently simulated even classically, which will provide valuable results for the future investigations on the genuine quantum hardwares \cite{HPWS18TNQML}.

\begin{figure}[tbp]
	\centering
	\includegraphics[angle=0,width=1\linewidth]{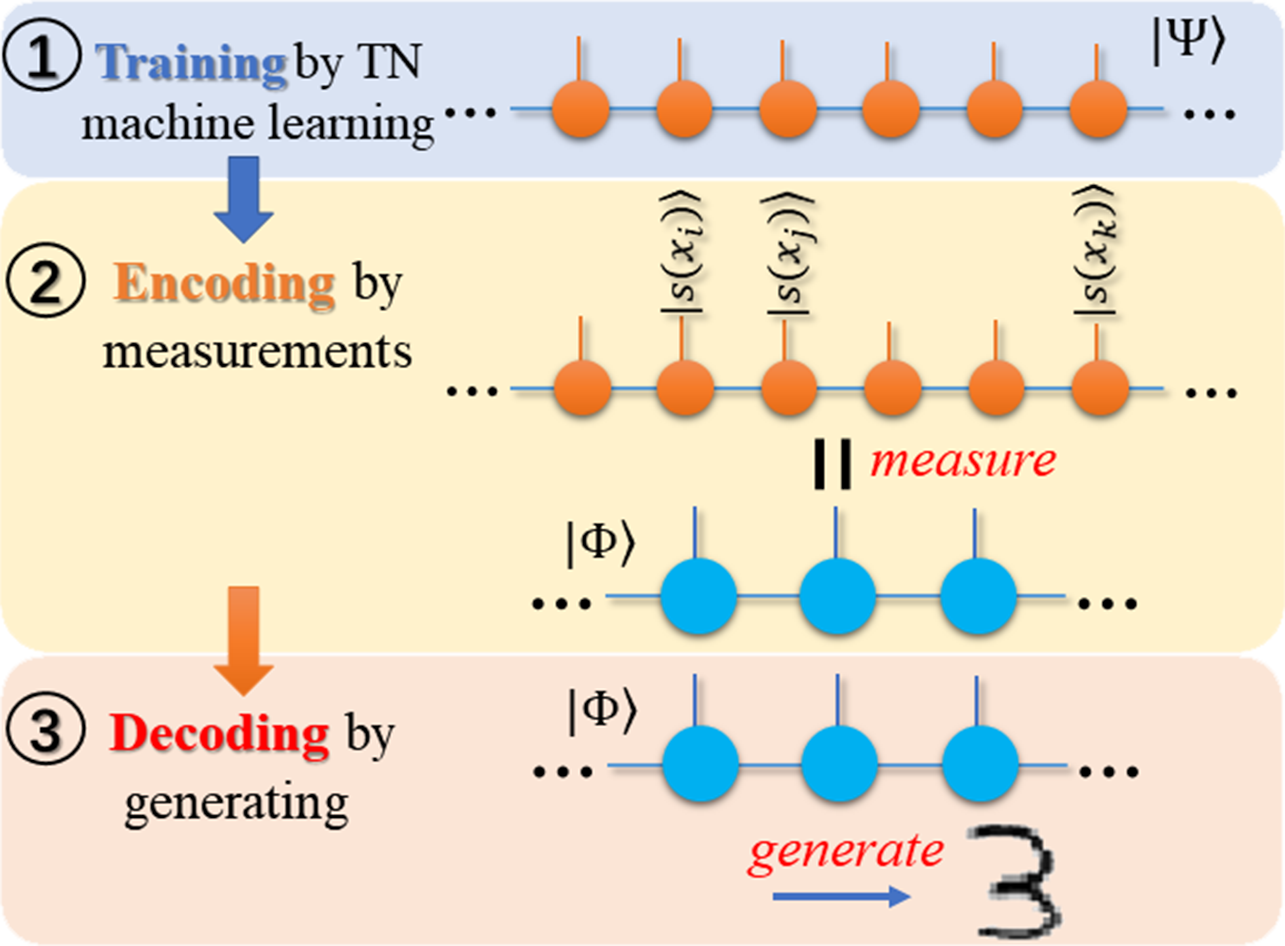}
	\caption{(Color online) Illustration of the main steps of TNCS: (1) train the Born machine $|\Psi \rangle$ representing the probability distribution of the data that Alice considers to send; (2) encode the specific piece of information to be sent by measuring $|\Psi \rangle$; (3) decode the information as a generative process by the measured Born machine.}
	\label{fig-TNCS}
\end{figure}

In this work, we propose tensor-network compressed sensing (TNCS), which permits efficient quantum communications of realistic data. The main idea is to encode and communicate the information by the measurements on the quantum state $|\Psi \rangle$ (also called Born machine \cite{cheng2018information}) trained by the unsupervised TN machine learning. To explain TNCS, let us consider the following scenario. Alice wants to send a piece of classical information $\{x\}$, e.g., an image of hand-written digit ``3'', to Bob in a secured way. She intends to send only a small number of pixels (or features in the terminology of machine learning) denoted by $\{x^{\text{sent}}\}$ to Bob by classical communication which might be unsafe or even public. The rest information $\{x^{\text{rest}}\}$ (with $\{x\} = \{x^{\text{sent}}\} \cup \{x^{\text{rest}}\}$) will be encoded in the Born machine $|\Psi \rangle$. To recover $\{x^{\text{rest}}\}$, Bob measures $|\Psi \rangle$ that is previously provided by Alice in the way determined by $\{x^{\text{sent}}\}$. After the measurements, $|\Psi \rangle$ will be projected to another entangled state denoted as $|\Phi\rangle$, and by design, $\{x^{\text{rest}}\}$ will be encoded in the separable state that has the minimal distance to $|\Phi\rangle$. Therefore, Bob can reliably recover $\{x^{\text{rest}}\}$ by measuring on $|\Phi\rangle$. A flowchart of TNCS is given in Fig. \ref{fig-TNCS}

There remain two key questions: how to construct $|\Psi \rangle$ and how to design the measurements on it, so that $\{x^{\text{rest}}\}$ can be encoded in $|\Phi\rangle$ in the above way. Our proposal is the following. First, Alice trains $|\Psi \rangle$ by the unsupervised TN machine learning algorithm \cite{HWFWZ17MPSML}, so that $|\Psi \rangle$ represents the probability distribution of a huge amount of information that Alice considers to send. $|\Psi \rangle$ is called a Born machine since the probability of each piece of information is the square of the corresponding coefficient in $|\Psi \rangle$ \cite{cheng2018information}. Then to send a specific piece of information, she chooses to send Bob the pixels, with which the uncertainty of the rest of the pixels in the probability distribution will be minimized. The full information $\{x\}$ is efficiently compressed to (or in other words, can be accurately reconstructed from) a small part of the image $\{x^{\text{rest}}\}$ and the Born machine $|\Psi \rangle$, similar to the (classical) compressed sensing schemes assisted by machine learning models, e.g., the auto-encoders \cite{Bora2017CSML, grover2018uncertainty}. 

We testify our TNCS with the datasets of hand-written digits and fashion images (namely MNIST \cite{MNIST} and fashion-MNIST \cite{fMNIST}). Any image in the training or testing sets can be reconstructed reliably and efficiently. The efficiency is indicated by the compression ratio $r = \#\{x^{[\text{sent}]}\} / \#\{x\} \simeq 10 \%$, where $\#\{x\}$ denotes the number of pixels in $\{x\}$. In other words, the information Bob accesses is about 10 times of the information that Alice needs to send through the classical channels. Most part of the information is encoded in the Born machine (quantum state), with the security guaranteed by the basic quantum principles. Similar to the compressed sensing, randomly choosing $\{x^{[\text{sent}]}\}$ already leads to small compression ratios. Better performance is reached by choosing $\{x^{[\text{sent}]}\}$ with a sampling protocol based on the entanglement of $|\Psi\rangle$, and by implementing post-selections to access $\{x^{[\text{rest}]}\}$. Finally, q-sparsity to characterize the sparsity of quantum state is proposed. For TNCS, q-sparsity characterizes how fast the Shannon entropy of the prbability distribution will decrease by measuring the Born machine $|\Psi\rangle$, and how efficient the compressed sampling can be via $|\Psi\rangle$. An empirical equation to estimate the required number of pixels for reliable reconstructions is given.

\section{Tensor-network compressed sensing}

Suppose Alice wants to send Bob an image of a hand-written digit ``3'' by TNCS (Fig. \ref{fig-TNCS}). She firstly trains the quantum state $|\Psi \rangle$ as the generative model for the training set of many ``3'' images in MNIST. This can be done with the unsupervised TN machine learning algorithm \cite{HWFWZ17MPSML}. The idea is to firstly map the images to quantum states. For example, the $n$-th each pixel ($0 \leq x_n \leq 1$) is mapped to a state of a qubit as
\begin{equation}\label{eq-featuremap}
x_n \to |s(x_n)\rangle = \cos (x_n\pi/2) |0\rangle + \sin (x_n\pi/2) |1\rangle,
\end{equation}
with $|0 \rangle$ and $|1 \rangle$ the two eigenstates of the Pauli matrix $\hat{\sigma}^z$. In this way, one image with pixels $\{x\} = (x_1, x_2, \cdots)$ is mapped to a separable state $|\psi \rangle = \prod_n |s(x_n)\rangle$. Then the Born machine $|\Psi \rangle$ is optimized to capture the probability distribution of the training set, by minimizing the distance (negative-log likelihood) to the probability distribution of the images in the training set. See more details in Appendix \ref{append-GTN}. Here, we take $|\Psi \rangle$ in the form of matrix product state (MPS) \cite{PVWC07MPSRev}. Note that TNCS is a general scheme, where one may also choose other TN forms to represent $|\Psi \rangle$, such as tree TN or MERA \cite{LRWP+17MLTN, HPWS18TNQML, CWXZ19generateTTNML}, or simply a quantum state without a specific entanglement structure.

In the sense of machine learning, though we only use the ``3'' images in the training set to optimize $|\Psi \rangle$, it is expected that $|\Psi \rangle$ approximately gives the probability distribution of any ``3'' images. In other words, $|\Psi \rangle$ learns the probability distribution of the ``3'' images from a finite (training) set, but can generalize to generate and/or recognize arbitrary ``3'' images that $|\Psi \rangle$ has never learned. The ability of a machine-learning model to process the information beyond the training set is known as the generalization power (see, e.g., \cite{PhysRevLett.59.2229}). As shown in the previous works \cite{SS16TNML, LRWP+17MLTN, HWFWZ17MPSML, GJLP18MPOML, GPC18gTNML, S18MERAML, HPWS18TNQML, CWXZ19generateTTNML}, TN models (including MPS) possess remarkable generalization power that is competitive to neural networks. Notably, TN models surpass neural networks as they allow to implement quantum process.

As $|\Psi \rangle$ gives the probability redistribution of the ``3'' images in the training set and beyond (due to its generalization power), it is then possible to use $|\Psi \rangle$ to communicate any ``3'' image. As a direct advantage, Alice can train $|\Psi \rangle$ without knowing the specific ``3'' image that will be sent to Bob. In other words, different ``3'' images can be communicated with the same state $|\Psi \rangle$, as long as $|\Psi \rangle$ can ``recognize'' (in the sense of machine learning) it as an image of ``3'' (see Appendix \ref{append-corr} for more discussions).

In the communication, Alice sends Bob only a small part of this image $\{x^{\text{sent}}\}$ and $|\Psi \rangle$; then Bob measures $|\Psi \rangle$ according to $\{x^{\text{sent}}\}$ as
\begin{equation}\label{eq-measure}
|\Phi\rangle = \prod_{x_{n} \in \{x^{[\text{sent}]}\}}  \langle s(x_{n}) |\Psi \rangle / C,
\end{equation}
with $C$ a constant to normalize $|\Phi\rangle$. $\{x^{\text{sent}}\}$ should be selected so that Bob can accurately reconstruct the rest of the pixels $\{x^{\text{rest}}\}$ from $|\Phi\rangle$. The selection of $\{x^{\text{sent}}\}$ is analog to the sampling process of compressed sensing. One may randomly choose $\{x^{\text{sent}}\}$ from $\{x\}$ (remind $\{x\} = \{x^{\text{sent}}\} \cup \{x^{\text{rest}}\}$). Each measurement by $|s(x_{n}) \rangle$ in Eq. (\ref{eq-measure}) is in fact a projection towards the separable state $|\psi \rangle = \prod_n |s(x_n)\rangle$. With sufficient $\{x^{\text{sent}}\}$, $|\Phi\rangle$ will Eventually be projected to such a state, where $\prod_n |s(x_n)\rangle$ ($x_n \in \{x^{\text{rest}}\}$) is the separable state that has the minimal distance to $|\Phi\rangle$ among all separable states. Therefore, Bob can access $\{x^{\text{rest}}\}$ by simply measuring on $|\Phi\rangle$. 

Let us consider that Bob only has one copy of $|\Psi\rangle$ (therefore only one copy of $|\Phi\rangle$), dubbed as one-shot measurement. To generate $\{x^{\text{rest}}\}$ from $|\Phi\rangle$, he measures the qubits in the basis of the Pauli matrix $\hat{\sigma}^z$. The probability $P(x_n)$ of the $n$-th pixel $x_n = 0$ or 1 is determined by $\hat{\rho}_n$ as $P(x_n) = \langle x|\hat{\rho}_n|x \rangle$ with $x = 0, 1$. 
$\hat{\rho}_n$ is the reduced density matrix with respect to the $n$-th qubit
\begin{equation}\label{eq-rhon}
\hat{\rho}_n = \text{Tr}_{/n} |\Psi \rangle \langle \Psi|,
\end{equation}
with $\text{Tr}_{/n}$ the trace over all degrees of freedom except for the $n$-th qubit. Note $\sum_x P(x) = \text{Tr} \hat{\rho}_n = 1$ due to the normalization of $|\Psi \rangle$. From the perspective of machine learning, such a way of obtaining $\{x^{\text{rest}}\}$ is in fact to generate $\{x^{\text{rest}}\}$ by the Born machine $|\Phi \rangle$ \cite{HWFWZ17MPSML}, and it is feasible in experiments. One drawback is that only black-or-white pixels ($x=0$ or 1) will be generated, not gray-scale ones. 

\begin{figure}[tbp]
	\centering
	\includegraphics[angle=0,width=1\linewidth]{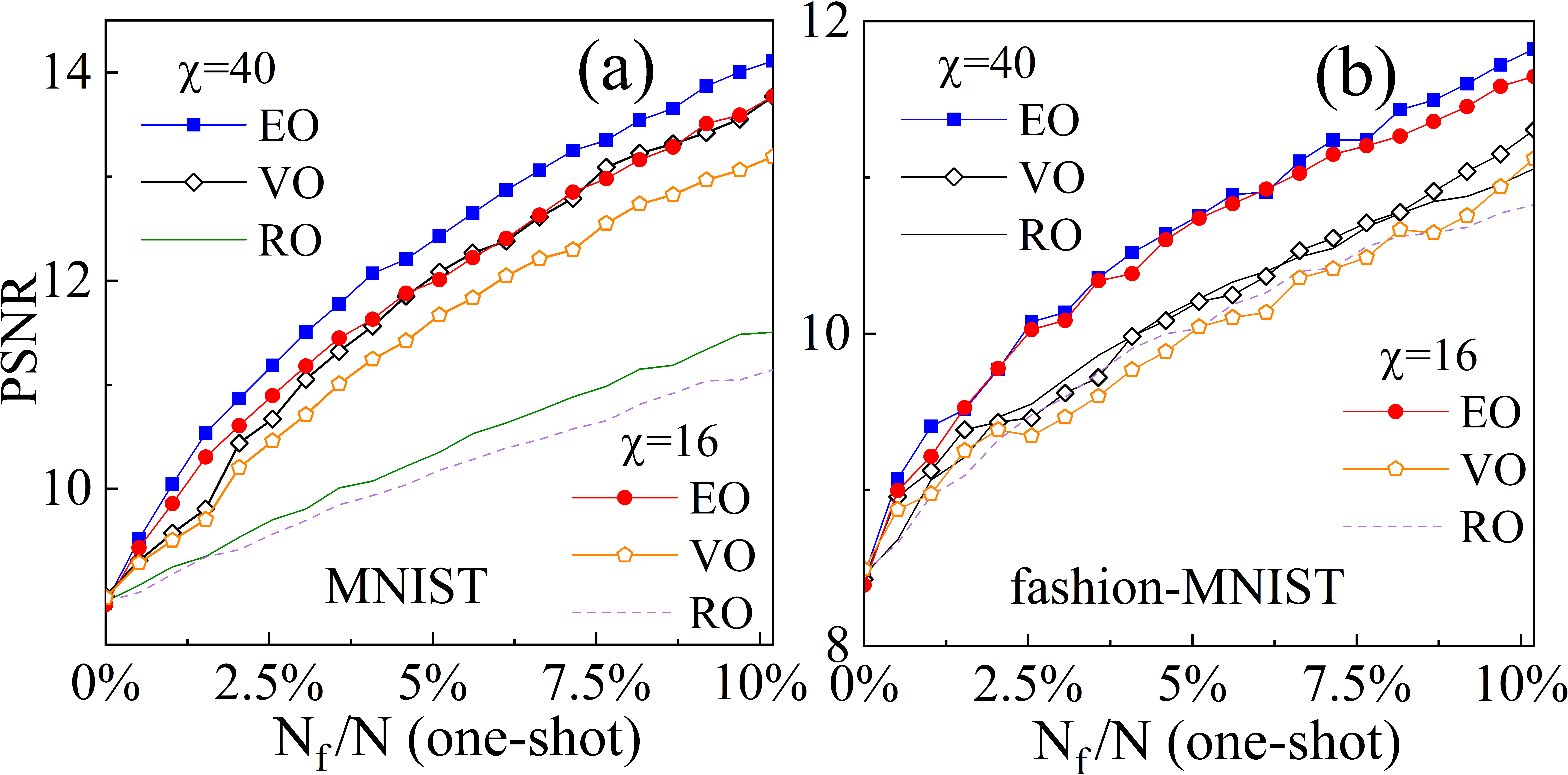}
	\includegraphics[angle=0,width=1\linewidth]{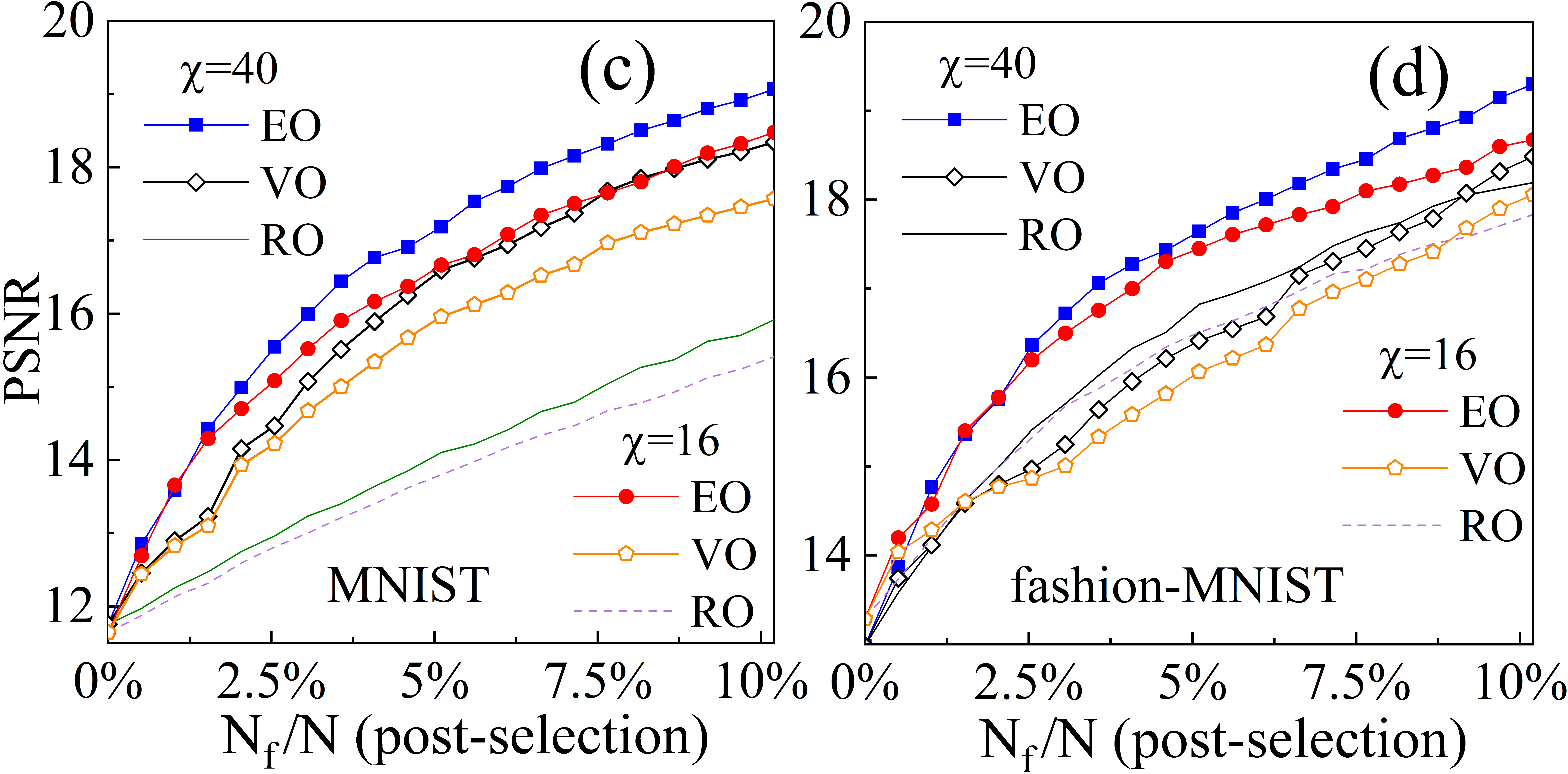}
	\caption{(Color online) Average peak signal-to-noise ratio (PSNR) of the constructed images in the testing dataset of the handwriting digits ``3'' in MNIST and the dresses in fashion-MNIST. The images are generated from $|\Phi \rangle$ in the one-shot way [(a) and (b)] or with the post-selection [(c) and (d)]. The dimension of the MPS is taken as $\chi=16$ or $40$. The number of known pixels for reconstruction ranges from about $N_f/N=0\%$ to $10\%$.}
	\label{fig-PSNR_Nf}
\end{figure}

We testify the TNCS with random selection and one-shot measurement on MNIST and fashion-MNIST datasets, which consists of realistic images of hand-written digits and Zalando's articles, respectively. Each dataset contains 10 classes of images, and in total has 60,000 training images and 10,000 testing images. Each image contains $28 \times 28 = 784$ gray-scale pixels. In Fig. \ref{fig-PSNR_Nf} (a) and (b), we show the accuracy of TNCS with different compression ratios $r=N_f/N$ (green solid and the purple dash lines). The accuracy is characterized by the average peak signal-to-noise ratio (PSNR), which (say between $\{x\}$ the reconstructed images $\{y\}$ ) is defined as
\begin{equation}\label{eq-PSNR}
\text{PSNR}(\{x\}, \{y\}) = 10 \log_{10} \frac{784}{\sum_n (x_n - y_n)^2}.
\end{equation}
We average the PSNR by the results of reconstructing all the images in the testing set, which the Born machine did not learn in the training process. We take the bond dimensions of the MPS $\chi=16$ and $40$. Generally, the PSNR increases with $r$ and $\chi$ as expected, and TNCS works well by simply sampling a small number of $\{x^{\text{sent}}\}$ randomly from $\{x\}$ and implementing one-shot measurement on $|\Psi\rangle$.

\section{Improving efficiency with entanglement-ordered sampling protocol and post-selections}

In the following, we propose to improve the performance (i.e., of higher PSNR and higher efficiency with smaller compression ratio) by incorporating with a sampling protocol based on entanglement and the post-selections of measurements.

Regarding the sampling, the results will change if Alice selects differently the $\{x^{\text{sent}}\}$. A natural selection way dubbed as variance ordering (VO) is to select the pixels according to the variance. The variance of the $n$-th pixel is calculated from the training set as
\begin{equation}\label{eq-variance}
V_n = \sum_i [x_{i, n} - (\sum_j x_{j, n} / K)]^2 / K.
\end{equation}
where $x_{i, n}$ is the $n$-th pixel in the $i$-th image of the training set and $K$ is the number of the training images. By choosing $\{x^{\text{sent}}\}$ as the pixels with the highest variance, the PSRN is obviously improved [see the black diamonds and orange pentagons in Fig. \ref{fig-PSNR_Nf} (a) and (b)]. 

A more reasoned way is to select based on the entanglement of $|\Psi\rangle$, so that $\{x^{\text{sent}}\}$ will minimize the uncertainty of $\{x^{\text{rest}}\}$ from the probability distribution given by the Born machine. Knowing $\{x^{[\text{sent}]}\}$, the (conditional) probability distribution of  $\{x^{[\text{rest}]}\}$ satisfies
\begin{equation}\label{eq-Px}
P(\{x^{[\text{rest}]}\} | \{x^{[\text{sent}]}\}) = |\prod_{x_{n} \in \{x^{[\text{rest}]}\}} \langle s(x_{n})| \Phi \rangle|^2,
\end{equation}
where $|s(x_{n})\rangle$ stands for the state associated with the $n$-th pixel $x_{n}$ [see Eq. (\ref{eq-featuremap})], and $|\Phi\rangle$ satisfies Eq. (\ref{eq-measure}). The task is to find the $N_f$ pixels $\{x^{[\text{sent}]}\}$ that minimize the Shannon entropy
\begin{eqnarray}\label{eq-Shannon}
S^{\text{Shan}} =&& -\sum_{\{x^{[\text{rest}]}\}} P(\{x^{[\text{rest}]}\} | \{x^{[\text{sent}]}\}) \nonumber \\&& \ln P(\{x^{[\text{rest}]}\} | \{x^{[\text{sent}]}\}).
\end{eqnarray}

Aiming at this task, let us begin with a simpler question: which pixel should be sent if Alice sends only one pixel? This can be determined by the single-site entanglement entropy (SEE) that (say for the $n$-th qubit) is defined as
\begin{equation}\label{eq-SEE}
S^{\text{ent}}_n = - \text{Tr} \hat{\rho}_n \ln \hat{\rho}_n.
\end{equation}
$S^{\text{ent}}_n$ quantifies the information of the rest of the system that will be gained if one has the information of the $n$-th qubit. Such a quantity has been utilized to safely reduce the number of pixels for efficient supervised TN machine learning \cite{LZLR18entTNML}. With $S^{\text{ent}}_n$, Alice can choose the $\tilde{n}$-th pixel with $\tilde{n} = \arg \max_{n} S^{\text{ent}}_n$, so that Bob will gain as much information as possible from one sent pixel.

Based on the above scheme, we propose the following Markov sampling strategy to select $\{x^{[\text{sent}]}\}$, dubbed as entanglement-ordered sampling protocol (EOSP).
\begin{enumerate}
	\item With an $N$-qubit state $|\Psi(N) \rangle$ (initialized as $|\Psi \rangle$), calculate the SEE $S^{\text{ent}}_n$ of all qubits, and find the qubit that has the maximal $S^{\text{ent}}_n$, i.e., $\tilde{n} = \arg \max_n S^{\text{ent}}_n$.
	
	\item From the reduced density matrix of the $\tilde{n}$-th qubit, $\hat{\rho}_{\tilde{n}}$, calculate its dominant eigenstate $|s_{\tilde{n}}\rangle$.
	
	\item Measuring the $\tilde{n}$-th qubit of $|\Psi(N) \rangle$, a $(N-1)$-qubit state is obtained as $|\Psi(N-1)\rangle = \langle s_{\tilde{n}}|\Psi(N) \rangle /C$, with $C$ a constant to normalize $|\Psi(N-1)\rangle$.
	
	\item If $N_f$ qubits have been measured, record the positions of these qubits, and transfer the pixels at these positions of the image to Bob. Note we have $|\Psi(N-N_f) \rangle = |\Phi \rangle$ [Eq. (\ref{eq-measure})]. Otherwise, go back to Step 1 and start again with $|\Psi(N-1) \rangle$.
\end{enumerate}

In short, EOSP selects the pixels in the order of entanglement (EO). A simple example that helps to understand the EOSP is provided in Appendix \ref{append-EOSP}. Similar strategies have been used in the classical compressed sensing. \cite{Bora2017CSML, grover2018uncertainty}, where the authors proposed to utilize the auto-encoders to significantly reduce the compression ratio. Note that these schemes are classical methods, where the security is not guaranteed by quantum physics. As shown in Fig. \ref{fig-PSNR_Nf} (a) and (b), EO achieves the highest PSNR among the three selection ways. Some discussions about the possible quantum advantages in the TNCS are given in Appendix \ref{append-quantum}.

Regarding the generation of $\{x^{[\text{rest}]}\}$ by the Born machine, we propose to use post-selections \cite{PhysRevLett.60.1351} to generate gray-scale images for higher accuracy (as the images in the datasets are gray-scale). We generate the pixels $\{x^{[\text{rest}]}\}$ by locating the separable state with maximal probability, i.e.,
\begin{equation}\label{eq-generatex}
\{x^{[\text{rest}]}\} = \arg\max_{\{x\}} |\prod_n \langle s(x_{n}) | \Phi \rangle|^2,
\end{equation}
where the product $\prod_n$ goes through $\{x^{[\text{rest}]}\}$. It means that each measurement basis $| s(x_{n}) \rangle$ is the dominant eigenstate of the corresponding single-site reduced density matrix of $|\Phi\rangle$ [Eq. (\ref{eq-rhon})]. Post-selections are required to realize such measurements. Fig. \ref{fig-PSNR_Nf} (c) and (d) show the results with post-selections. One can see that the PSNR's for all three selection ways (EO, VO, and RO) are significantly improved. With EO and post-selections, the image can be accurately and efficiently communicated to Bob (with PSNR$\simeq 20$) for $r\simeq 10 \%$. Same as the existing quantum communication schemes, the security of the communications with TNCS are guaranteed by the fundamental principles of quantum physics (more discussions are given in Appendix \ref{append-QCM}).


\section{Q-sparsity}

\begin{figure}[tbp]
	\centering
	\includegraphics[angle=0,width=0.9\linewidth]{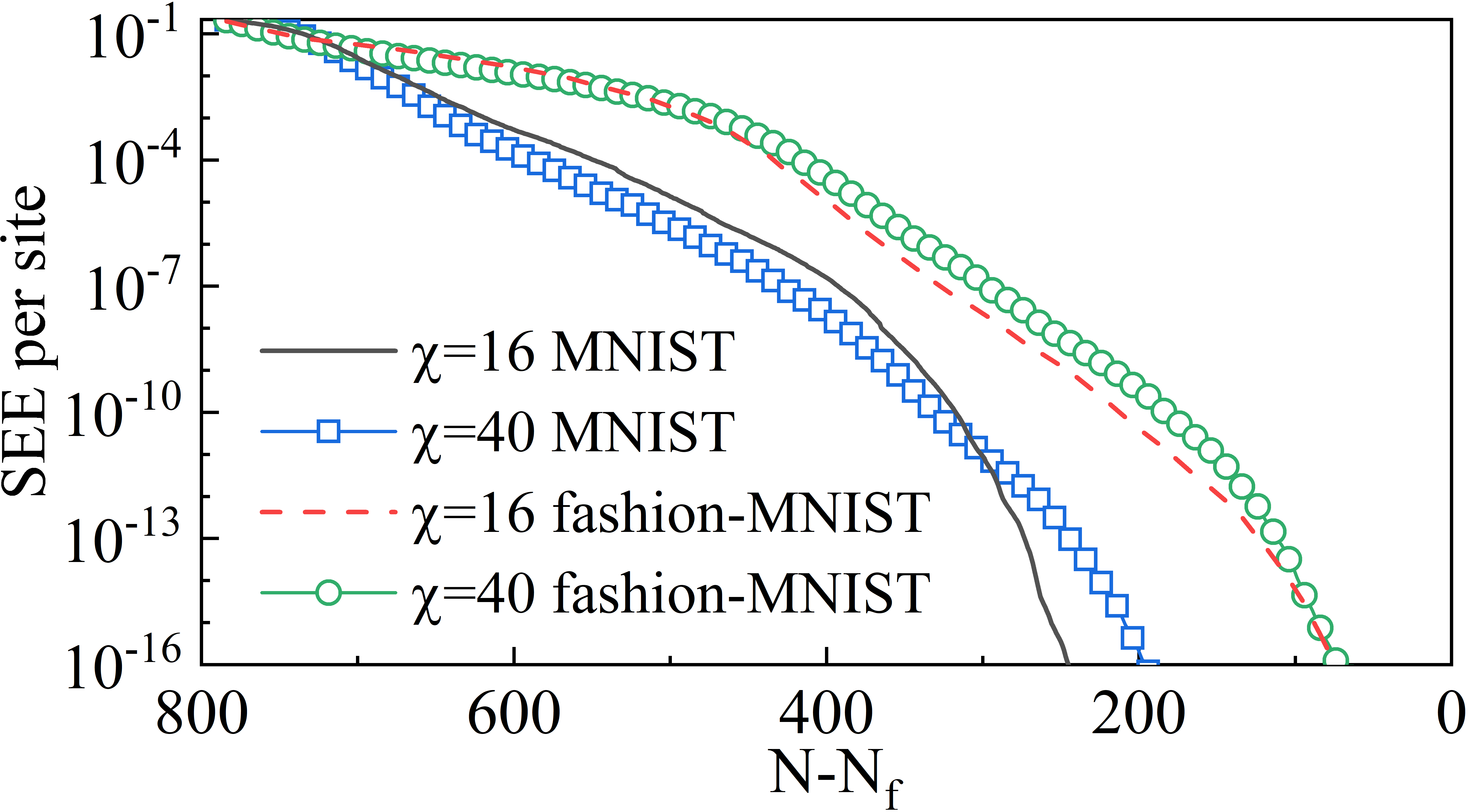}
	\caption{(Color online) SEE [Eq. (\ref{eq-SEE})] per site of $|\Phi \rangle$ in EOSP versus the number of the unmeasured qubits $N-N_f$. The more steeply the SEE per site decays, the faster the information of a quantum state can be gained by measurements.}
	\label{fig-see_av}
\end{figure}

A prerequisite for the conventional compressed sensing to work is the sparsity of the signals. For processing images, it is known that the signals are usually not sparse in the real space. Therefore, transformation (such as discrete cosine/wavelet transformation) is implemented to transform to another space in which the signals are sparse.

In TNCS, sparsity is gained in a completely different way, which is by mapping the data to the higher-dimensional quantum Hilbert space. This is analog to the support vector machines \cite{CV95SVM} by mapping to a higher-dimensional space where the data can be better classified. In the unsupervised TN machine learning algorithm, each pixel $x$ is mapped to the state of a qubit [Eq. (\ref{eq-featuremap})], then one image is mapped to the direct product state of $N$ qubits with $N$ the number of pixels. Such a vector is defined in a $(2^N)$-dimensional space $\mathcal{H}$. The MPS $|\Psi \rangle$ describes the joint probability distribution of the ``vectorized'' images in $\mathcal{H}$. Essential, one still deals with the data in the real space. However, the probability distribution becomes sparse in this higher-dimensional real space, since it can be well captured by an MPS. An MPS is sparse because such a representation can only reach a small corner of $\mathcal{H}$ that satisfies the so-call 1D area law of entanglement entropy \cite{ECP10AreaLawRev, F13arealawTRG}.

However, it is not easy to characterize the sparsity of an MPS, as its dimension is exponentially large. We here propose to use EOSP to do so. In each step of EOSP, the qubit with the maximal SEE is measured. The entanglement of the state $|\Phi \rangle$ formed by the unmeasured qubits decreases after each measurement. Fig. \ref{fig-see_av} shows the SEE per site $\bar{S}(\tilde{n}) = \sum_n S^{\text{ent}}_n(\tilde{n}) / \tilde{n}$ of $|\Phi(\tilde{n}) \rangle$ [see Eq. (\ref{eq-SEE})] with different number of unmeasured qubits $\tilde{n}$. One can see that $\bar{S}(\tilde{n})$ decays rapidly with $N-\tilde{n}$, meaning the unmeasured qubits are almost in a separable state for small $\tilde{n}$. For $\bar{S}(\tilde{n})=0$, no information will be gained by knowing the unmeasured pixels. It means all information is contained in the measured pixels, and there is no uncertainty for the rest pixels, when $\bar{S}(\tilde{n})$ becomes zero.

From the implication of $\bar{S}(\tilde{n})$ discussed above, we define q-sparsity to qualitatively describe the sparsity of a quantum state (including MPS) as
\begin{equation}\label{eq-sparsity}
\mathcal{S}^q = \prod_{\tilde{n}=1}^{N} d^{\frac{\bar{S}(\tilde{n})}{\ln d} - 1},
\end{equation}
with $d$ the dimension of one vectorized pixel. For qubits, we have $d=2$. The Q-sparsity characterizes how fast the information of a quantum state can be gained (or how fast the uncertainty of the rest can be reduced) by measurements. Take the $N$-qubit GHZ state as an example. We have $\bar{S}(N) = \ln 2$ originally, and $\bar{S}(\tilde{n} \neq N) = 0$ after one measurement. Therefore, we have $\mathcal{S}^q = 2^{-N+1}$. For the conventional $k$-sparsity, we have $\mathcal{S}^k = 2/2^N = 2^{-N+1}=\mathcal{S}^q$ since it only has two non-zero coefficients in the $2^N$-component vector. Take the maximally-entangled state \cite{GB98maxent} as another example. We have $\mathcal{S}^q=1$ since $\bar{S}(\tilde{n}) = \ln 2$ for any $\tilde{n}$. For the generative MPS's, we numerically have $\mathcal{S}^q = 2^{-768.6}$ and $2^{-765.6}$ with $\chi=16$ for MNIST and fashion-MNIST, respectively, and $\mathcal{S}^q = 2^{-770.0}$ and $2^{-767.5}$ with $\chi=40$.

For TNCS, $\mathcal{S}^q$ characterizes the efficiency, i.e., the compression ratio. The smaller $\mathcal{S}^q$ is, the faster $\bar{S}(\tilde{n})$ decays in general with the measurements, and the less $\{x^{[\text{sent}]}\}$ Bob will require to accurately reconstruct the full information by TNCS. Therefore, analog to the conventional compressed sensing, TNCS requires the probability distribution to be sparse in the higher-dimensional Hilbert space, i.e., $N + \log_2 \mathcal{S}^q \ll N$. Based on our results, the required number of $\{x^{[\text{sent}]}\}$ to reach $\text{PSNR} \simeq 20$ can be estimated as
\begin{equation}\label{eq-nf}
N_f \simeq c (N+\log_2 \mathcal{S}^q),
\end{equation}
with $c\simeq 6$ for both MNIST and fashion-MNIST.

\section{Summary}
In this work, we propose a quantum compressed sensing approach by combining the ideas of compressed sensing, quantum communication, and unsupervised TN machine learning. The key step is to train the quantum state $|\Psi \rangle$ (a Born machine) by the unsupervised TN machine learning algorithm, so that the targeted piece of information can encoded in the separable state with the minimal distance to $|\Phi \rangle$ that is obtained by measuring on $|\Psi \rangle$ in a designed way. The q-sparsity is proposed as a fundamental property of quantum states, and is used to estimate the efficiency of TNCS. We apply TNCS to the realistic datasets (hand-written digits and fashion images). Unique advantages of TNCS are demonstrated, where images can be compressed and transferred in a comparable efficiency and accuracy than the classical methods, and at the same time the security of the communications is guaranteed by the fundamental quantum principles.

\section*{Acknowledgments}
SJR is grateful to Ding Liu for helpful discussions. This work was supported by Beijing Natural Science Foundation (No. 1192005 and No. Z180013), National Natural Science Foundation of China (No. 11675113) and Beijing Municipal Commission of Education (KZ201810028042). ML acknowledges the Spanish Ministry MINECO (National Plan 15 Grant: FISICATEAMO No. FIS2016-79508-P, SEVERO OCHOA No. SEV-2015-0522, FPI), European Social Fund, Fundació Cellex, Generalitat de Catalunya (AGAUR Grant No. 2017 SGR 1341 and CERCA/Program), ERC AdG OSYRIS and NOQIA, EU FETPRO QUIC, and the National Science Centre, Poland-Symfonia Grant No. 2016/20/W/ST4/00314. ZZS and GS are supported in part by the NSFC (Grant No. 11834014), the National Key R\&D Program of China (Grant No. 2018FYA0305800), the Strategic Priority Research Program of CAS (Grant No. XDB28000000), and Beijing Municipal Science and Technology Commission (Grant No. Z118100004218001).

\setcounter{equation}{0}
	\setcounter{figure}{0}
	\setcounter{table}{0}
	\makeatletter
	\renewcommand{\theequation}{A\arabic{equation}}
	\renewcommand{\thefigure}{A\arabic{figure}}
	\renewcommand{\thetable}{A\arabic{table}}

\appendix


\section{Unsupervised tensor-network machine learning algorithm}
\label{append-GTN}
In the generative TN machine learning algorithm proposed in Ref. \cite{HWFWZ17MPSML}, each image is mapped to a product state of $N$ qubits as $|\phi_i \rangle = \prod_{n} |s(x_{i, n}) \rangle$ with $|s(x_{i, n}) \rangle = \cos (x_{i,n}\pi/2) |0\rangle + \sin (x_{i,n}\pi/2) |1\rangle$ and $N$ the total number of pixels in one image. Here, $x_{i,n}$ is the $n$-th pixel (gray with $0\leq x_{i,n} \leq 1$) of the $i$-th image. The coefficients in the quantum state $|\Psi\rangle$ are optimized to minimize the negative log-likelihood (NLL) defined as
\begin{equation}\label{eq-NLL}
f = \ln |\langle \Psi | \Psi \rangle|^2 - \frac{\sum_i \ln |\langle \Psi | \phi_i \rangle|^2}{N}.
\end{equation}
The summation $\sum_i$ is over all the training images. NLL characterizes the resemblance between two probability distributions.

In this work, we choose the TN to be matrix product state (MPS). The coefficients of $|\Psi \rangle$ are in a special form satisfying
\begin{equation}\label{eq-MPS}
|\Psi \rangle = \sum_{\{a\}}  \prod_{n} \sum_{s_n=0,1} A^{[n]}_{s_n a_n, a_{n+1}} |s_n\rangle.
\end{equation}
$A^{[n]}$ represents a tensor that corresponds to the $n$-th pixel. The indexes $\{a\}$ are known as virtual bonds of the MPS; their dimensions are bounded by $\dim(a_n) \leq \chi$, with $\chi$ called virtual bond dimension. MPS is an efficient representation of quantum-many-body states where the total number of parameter scales linearly  with $N$ as $\sim 2N\chi^2$. Note that the dimension of the Hilbert space actually scales exponentially as $\sim 2^N$. The tensors in the MPS are updated alternatively by the gradient method as $A^{[n]} \leftarrow A^{[n]} - \tau \partial f/ \partial A^{[n]}$, with  $\tau$ the gradient step; see Ref. \cite{SS16TNML} or \cite{HWFWZ17MPSML} for more details. 

After converging, $|\Psi \rangle$ gives the joint probability of the pixels. The probability for any image $\{x\}$ in $|\Psi \rangle$ is given as 
\begin{equation}\label{eq-prob}
P(\{x\}) = |\prod_{n} \langle s(x_{n}) | \Psi \rangle|^2.
\end{equation}
Note the probability is the square of the corresponding coefficient, thus such a TN state is also called the Born machine \cite{cheng2018information}.

\section{Ambiguous correlations of information in TNCS}
\label{append-corr}
Another immediate question about TNCS is how to determine the samples (denoted by $\mathbb{A}$) for training the Born machine $|\Psi \rangle$, and what are the relations to the information (denoted by $\mathbb{B}$) that can be transferred or reconstructed through $|\Psi \rangle$. Obviously, we have $\mathbb{A} \subseteq \mathbb{B}$. The size of the complementary set $\mathbb{C} = \mathbb{B} - \mathbb{A}$ characterizes the generalization power of the Born machine.

Evidently, $\mathbb{C}$ has to be ``ambiguously'' correlated to $\mathbb{A}$ somehow. Let us consider an extreme situation, where all training samples in $\mathbb{A}$ are formed by uncorrelated random numbers. The trained state $|\Psi \rangle$ is an entangled state. However, such a state obviously cannot be used to effectively transfer a random image as no correlations exist between the random image and the state.

In this work, we choose $\mathbb{A}$ and $\mathbb{B}$ as the training and testing images of the same dataset, respectively. For instance, $\mathbb{A}$ and $\mathbb{B}$ are handwritten digits ``3'' or images of dresses. Although the ``microscopic information'' (pixels) of all the images in $\mathbb{A}$ and $\mathbb{B}$ are different from each other, a human being can recognize the ``macroscopic information'' of each image as a digit ``3'' (or a dress) without any problem. This suggests that $\mathbb{A}$ and $\mathbb{B}$ (thus $\mathbb{A}$ and $\mathbb{C}$) must be correlated somehow. In other words, we here ensure the existence of the ``ambiguous'' correlations between $\mathbb{A}$ and $\mathbb{B}$ by the ``macroscopic'' information. 

With the TN machine learning, we can define the ``ambiguously'' correlation in a relatively more rigorous way: $\mathbb{A}$ and $\mathbb{B}$ are ``ambiguously'' correlated if the Born machine trained by $\mathbb{A}$ can accurately recognize the data in $\mathbb{B}$. For instance, one may train two Born machines by the ``3'' and ``4'' images in the training set, respectively, and construct a classifier that accurately recognizes ``3'' and ``4'' images \cite{SPLRS19GTNC}. To classify an image in $\mathbb{B}$ or the testing set, one compares the probability of have this image in the two Born machines, and classification is given by finding the largest probability.

The above recognition scheme can give us many useful information. For instance, the Born machine trained by the ``3'' images can be used to implement the TNCS for an image ``3'' written by the reader, as long as it can be recognized by the Born machine. Obviously, the TNCS cannot be implemented by the Born machine of ``3'' if the reader writes a ``4''. How to more rigorously characterize and quantify such ambiguous correlations is an important issue to TNCS. One direction is to develop more universal classifiers for pattern recognition (not limited to digits or some certain kind of data). This will also be helpful to further understand and model the recognition process.

\section{A simple example to understand entanglement-ordered sampling protocol}
\label{append-EOSP}
To explain why the entanglement-ordered sampling protocol (EOSP) works, let us consider the following four-qubit state as an example,
\begin{eqnarray}\label{eq-Apsy4}
|\Psi \rangle &=& (\frac{\sqrt{2}}{2} |01\rangle + \frac{\sqrt{2}}{2} |10\rangle) \otimes (\frac{1}{2} |01\rangle + \frac{\sqrt{3}}{2} |10\rangle) \nonumber \\ &=&  \frac{\sqrt{2}}{4} |0101\rangle + \frac{\sqrt{6}}{4} |0110\rangle + \frac{\sqrt{2}}{4} |1001\rangle + \frac{\sqrt{6}}{4} |1010\rangle. \nonumber \\
\end{eqnarray}
Such a state can describe a dataset of four images $(0, 1, 0, 1)$, $(0, 1, 1, 0)$, $(1, 0, 0, 1)$, and $(1, 0, 1, 0)$, with the probability $P = 1/8$, $3/8$, $1/8$, and $3/8$, respectively.

If Alice wants to send two pixels and encode the rest two in the state, the pixel that Alice should firstly choose is obviously the first (or the second) pixel. Since the first two qubits are in the maximally entangled state, one of the pixels can be determined by knowing the other pixel. The second pixel Alice chooses should be the third or the forth one. These two qubits are entangled (but not maximally), thus knowing one of them will gain certain (but not the full) information of the other. In all, Alice should send the first (or second) and the third (or the forth) pixels to Bob.

The EOSP gives the same answer. The SEE of $|\psi \rangle$ satisfies $S^{\text{ent}}_1 = S^{\text{ent}}_2 = \ln 2 \simeq 0.693$, and $S^{\text{ent}}_3 = S^{\text{ent}}_4 = -\frac{1}{4} \ln \frac{1}{4} - \frac{3}{4} \ln \frac{3}{4} \simeq 0.562$. In the step 1 of the EOSP, Alice chooses the first or the second pixel. The reduced density matrices satisfy $\hat{\rho}_1 = \hat{\rho}_2 = I / 2$, with $I$ the $2 \times 2$ identity. Therefore, Alice decides to measure the first qubit by $|0\rangle \langle 0|$ or $|1\rangle \langle 1|$. In either case, the resulting three-qubit state will be $|\Psi(3) \rangle = |x\rangle \otimes (\frac{1}{2} |01\rangle + \frac{\sqrt{3}}{2} |10\rangle)$ with $x=0$ or $1$. In the second iteration, Alice has $S^{\text{ent}}_2 = 0$ and $S^{\text{ent}}_3 = S^{\text{ent}}_4 \simeq 0.562$, thus she decides to send the third (or forth) pixel. In comparison, Alice will choose to send the first and second pixels according to the variance, which is not a good idea since Bob will not be able to gain any information about the third and forth pixels. Again, we would like to emphasize that this example is to help understand EOSP; it is too simple to draw any general conclusions about the advantages/disadvantages of quantum methods over classical ones.

\section{Quantum nature in TNCS}
\label{append-quantum}
\begin{figure}[tbp]
	\centering
	\includegraphics[angle=0,width=1\linewidth]{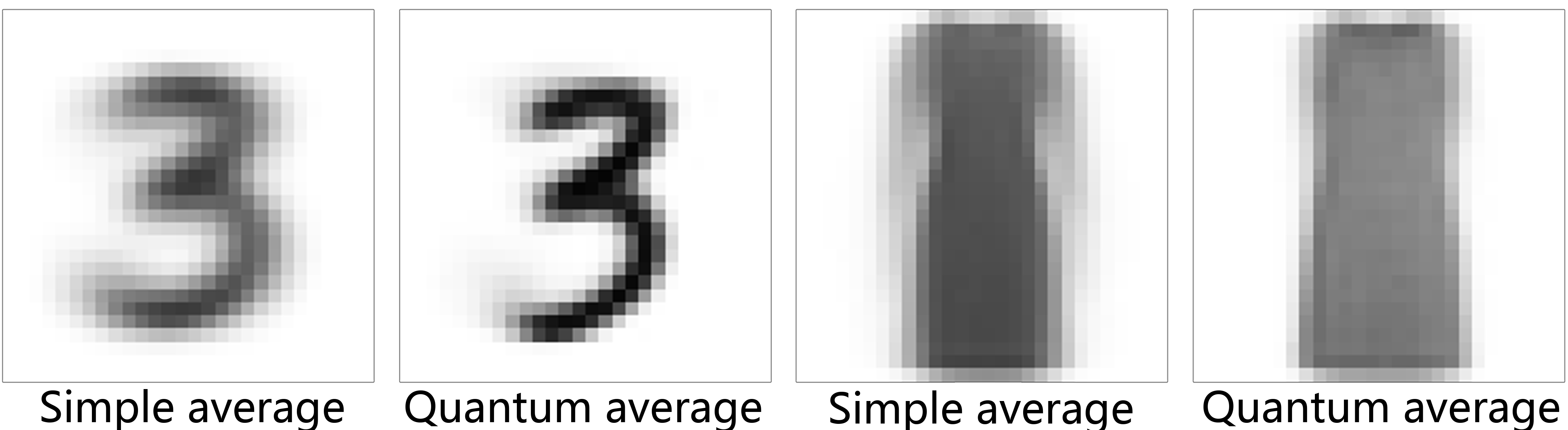}
	\caption{The images by taking simple average of each pixel and by taking the quantum average (generated by MPS with no known pixel).}
	\label{fig-Average}
\end{figure}

With $N_f = \#\{x^{\text{sent}}\} = 0$, Bob will randomly generate an image according to the probability distribution give by $|\Psi \rangle$. If the post-selections are used, the result will approach to the separable state that has the minimal distance to $|\Psi \rangle$. This separable state gives the image that has the maximal probability in the probability distribution. We dub such an image from no known pixel as the quantum average. One $|\Psi\rangle$ gives one unique quantum average (we assume that all $\hat{\rho}_n$'s have non-degenerated eigenvalues). As shown in Fig. \ref{fig-Average}, the quantum average is different from the simple average $\bar{x}_n = \sum_i x_{i,n} /K$, since no correlations are considered in the simple average. Correlations (and entanglement) are considered in the quantum average when calculating the reduced density matrix.

\begin{figure}[tbp]
	\centering
	\includegraphics[angle=0,width=1\linewidth]{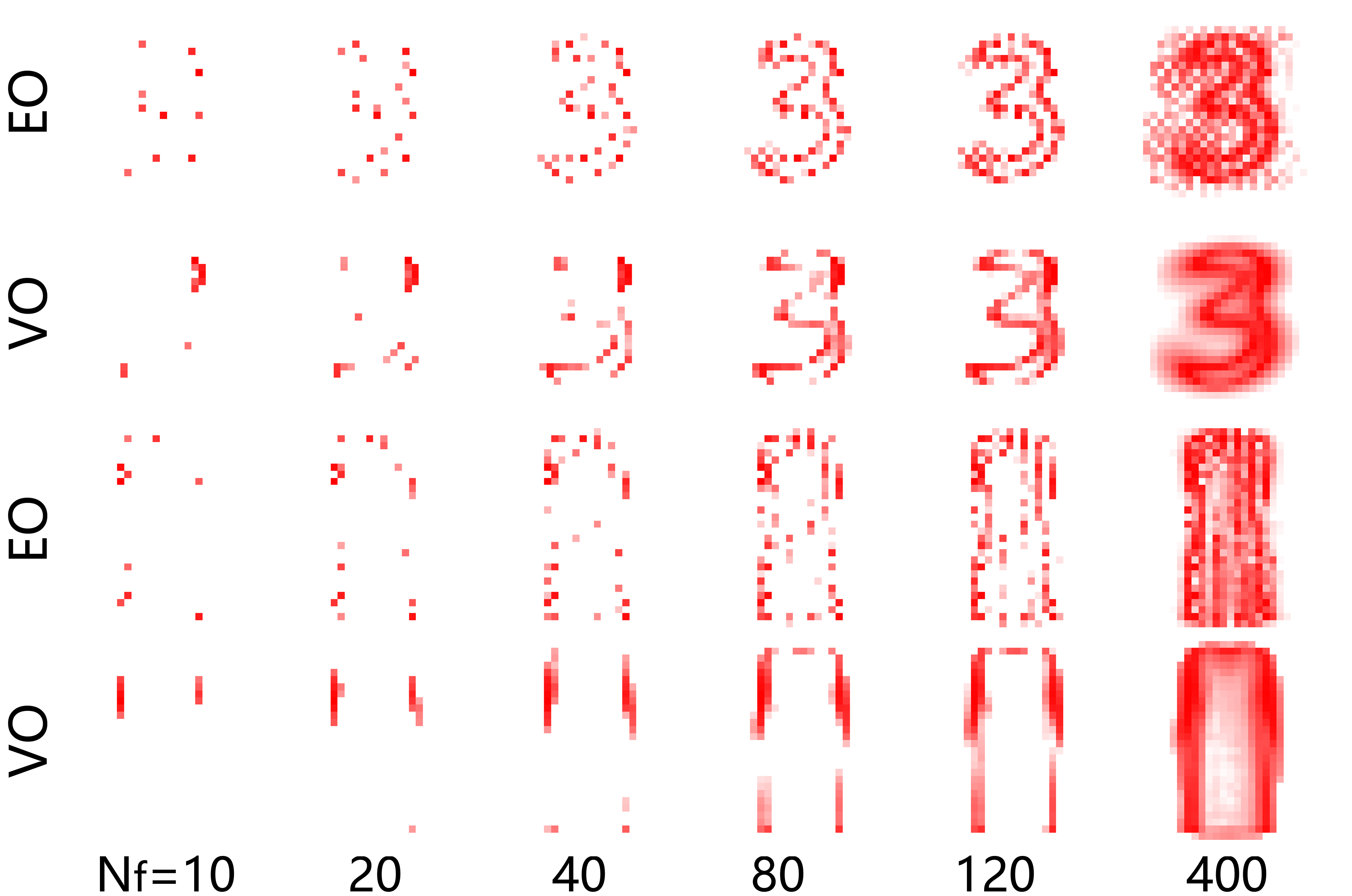}
	\caption{(Color online) Which $N_f$ pixels are selected in EO and VO. To illustrate the orders by color, we mark a pixel redder than those behind this pixel in the order.}
	\label{fig-OrderMap}
\end{figure}

Fig. \ref{fig-OrderMap} shows which pixels are selected in EO and VO with different values of $N_f$. To illustrate the orders, we mark a pixel redder than those pixels that are behind this pixel in the order. Both EO and VO manage to capture the general shapes. Particularly, the ``checker-board'' pattern appears in EO with relatively large $N_f$. This brings higher efficiency for the following reason. Since each two nearest-neighbor pixels should possess a strong correlation, the corresponding qubits are expected in a highly entangled state. It means that one only needs to know the information of one qubit (pixel) to access the information of the other qubit (pixel). Taking the maximally entangled two-qubit state $|01\rangle + |10\rangle$ as an example, if one knows that the first qubit is in the state $|0\rangle$ (or $|1\rangle$), meaning that the first pixel $x_1=0$ (or $x_1=1$), one will know that the second qubit is in the state $|1\rangle$ (or $|0\rangle$), meaning that the second pixel $x_2=1$ (or $x_2=0$). In this case, one only needs to send the information of one of the pixels, and the rest will be obtained from the state.

Intuitively, both the quantum entanglement and the (classical) variance measure the amount of the carried information. For instance, considering a pixel (labeled as $n$) that is always black in all the training images, such a pixel obviously carries no information, and we have $S_n = V_n = 0$. On the other hand, if a pixel changes dramatically with the training images, not necessarily but normally, this pixel may contain more information, and we will have large $S_n$ and $V_n$. One essential difference is that $S_n$ and $V_n$ are properties from the quantum state and the classical data, respectively. In our case, the quantum quantity (EO) outperforms the classical one (VO), providing an evidence of the quantum advantage in the TNCS.

However, we cannot stating here the general quantum advantages over classical information with these two specific methods. As we stated before, EO considers certain non-local properties while VO is purely local. Nevertheless, TNCS indeed provides a new path to investigate quantum advantages over classical information techniques. Several important and interesting questions are to be investigated, such as how to define new (classical or quantum) quantities that better suppress the compression ratio and/or increase the accuracy. Possible choices include the (classical) co-variance of the training data, the (quantum) correlation functions from $|\Psi \rangle$, and the multipartite entanglement. The performance of both quantum and classical methods for selecting $\{x^{\text{sent}}\}$ need to be pushed to their limits to discuss more clearly about the possible quantum advantages.

\section{TNCS and quantum encrypted communication}
\label{append-QCM}
In the scenario depicted above, TNCS can be used to securely send information via quantum states. Since $|\Psi\rangle$ cannot be cloned, the information is secured under the assumption that those without $|\Psi \rangle$ cannot reconstruct the full information solely from $N_f \ll N$ pixels. Moreover, there are many ways to enhance the security to avoid that the full information be cracked from the known pixels.

For example, Alice can introduce a one-to-one (reversible) deterministic map $\{y^{[\text{sent}]}\} = F(\{x^{[\text{sent}]}\}; \{x^{[\text{rest}]}\})$ to encrypt $\{x^{[\text{sent}]}\}$. Without $F$, the $\{x^{[\text{sent}]}\}$, which might be unsafe, could contain critical information (see for example Fig. \ref{fig-OrderMap}, which are almost meaningful images for $N_f>40$). The purpose of $F$ is to avoid containing any meaningful information in $\{x^{[\text{sent}]}\}$.

Such a $F$-encrypted TNCS will contain the following steps: 1) Alice designs the function $F$, and trains $|\Psi \rangle$ by the images formed by $\{x^{[\text{rest}]}\}$ and $\{y^{[\text{sent}]}\}$; 2) Alice sends $|\Psi \rangle$ to Bob; 3) For the information to be sent, Alice sends $\{y^{[\text{sent}]}\} = F(\{x^{[\text{sent}]}\}; \{x^{[\text{rest}]}\})$ and the function $F$ to Bob through classical channels that may not be safe; 4) Bob obtains $\{x^{[\text{rest}]}\}$ by $|\Psi \rangle$ and $\{y^{[\text{sent}]}\}$ (same to the standard TNCS), and obtains $\{x^{[\text{sent}]}\}$ by $\{y^{[\text{sent}]}\}$, $\{x^{[\text{rest}]}\}$, and the inverse of $F$. Then Bob will have the full information $\{x^{[\text{sent}]}\} + \{x^{[\text{rest}]}\}$. The information will be safe since those without $|\Psi\rangle$ cannot have $\{x^{[\text{rest}]}\}$, thus cannot obtain $\{x^{[\text{rest}]}\})$ even if they have $F$ and $\{y^{[\text{sent}]}\}$.

Since the information to be sent is not restricted to the data that train $|\Psi \rangle$, Alice can provide previously the copies of $|\Psi \rangle$ to multiple parties, and send any piece of ``ambiguously'' correlated information to each party anytime afterwards. Different pieces of information can be sent via the copies of the same state.

Meanwhile, Alice does not allow other parties to access the coefficients of $|\Psi \rangle$, to guarantee herself as the only provider of the state. One potential risk is that Alice provides too many copies of $|\Psi \rangle$ to others, with which the coefficients of $|\Psi \rangle$ can be cracked by, e.g., quantum state tomography \cite{VR89tomo}. In our case, this risk is low since $N$ is large, and it can be easily controlled by the number of the states provided to other parties.

In the scenario discussed above, Alice sends a small part of the classical information $\{x^{[\text{sent}]}\}$ and the whole state $|\Psi \rangle$ to Bob. Bob then generates the missing information $\{x^{[\text{rest}]}\}$ from $|\Psi \rangle$ and $\{x^{[\text{sent}]}\}$. In this scenario, one does not need stabilize remote entanglement between qubits that are far separately. 

This process can be replaced by a more standard quantum communication scheme. First, Alice trains and prepares $|\Psi \rangle$. Then she sends the qubits corresponding to $\{x^{[\text{rest}]}\}$ to Bob, and keeps those corresponding to $\{x^{[\text{sent}]}\}$ to herself. Note that these qubits of $\{x^{[\text{sent}]}\}$ and $\{x^{[\text{rest}]}\}$ form the whole entangled state $|\Psi \rangle$. To send the information, Alice measures her qubits according to $\{x^{[\text{sent}]}\}$. Afterwards, Bob generates the $\{x^{[\text{rest}]}\}$ from his qubits. 

In this scenario, Alice only gives a part of the qubits in $|\Psi \rangle$ to Bob or other receivers, and does not need to transfer the information of $\{x^{[\text{sent}]}\}$ through classical channel. It avoids the risks in communicating $\{x^{[\text{sent}]}\}$ classically. The disadvantage is that the qubits with Alice and the receivers need to be kept remotely entangled until Alice implements the measurement on her qubits.

\section{More numerical data of TNCS}
\label{append_moredata}

\begin{figure*}[tbp]
	\centering
	\includegraphics[angle=0,width=0.9\linewidth]{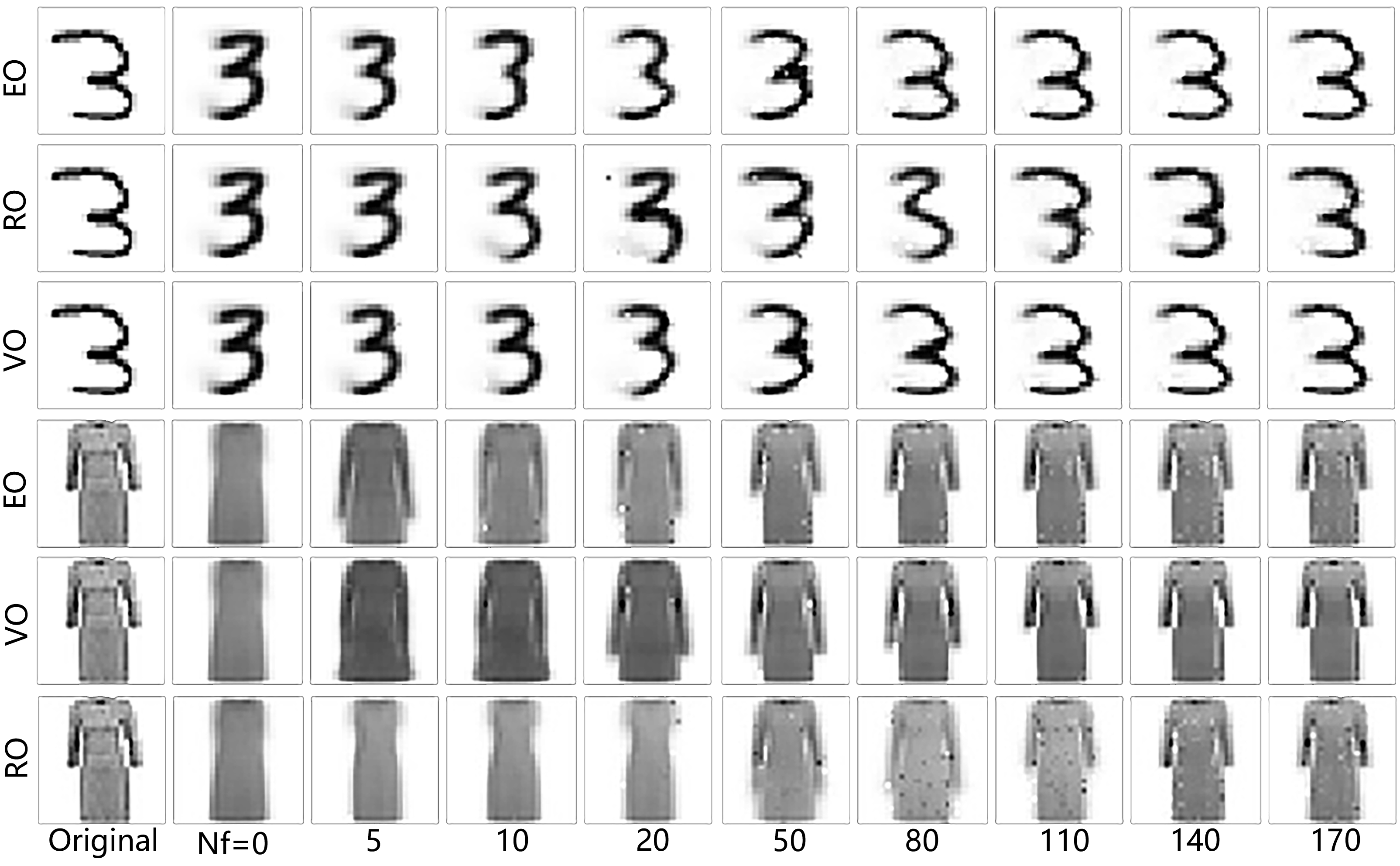}
	\caption{Examples of original and generated images in MNIST and fashion-MNIST in the entanglement order (EO), random order (RO), and variance order (VO). The number of known features $N_f$ varies from 0 to 170, while the total number of features in an image is 784. We take the bond dimension of the generative MPS as $\chi=40$.}
	\label{fig-VaryNumFeature}
\end{figure*}

For $N_f=0$, Bob generates the image that has the maximal probability in $|\Psi\rangle$, namely the quantum average. Thus, Alice needs to send $N_f>0$ pixels, with which Bob can implement the measurements accordingly so that $|\Psi\rangle$ will be projected to have $\{\tilde{x}\}$ as the configuration with the maximal probability. For $N_f>0$, the more known pixels there are, the more accurately $\{x\}$ will be encoded in the measured state. Fig. \ref{fig-VaryNumFeature} demonstrates two original images and the reconstructed images with different numbers of known pixels $N_f$ picked in three different orders (EO, RO, and VO). Take the reconstruction of a dress image as an example (last three rows in Fig. \ref{fig-VaryNumFeature}). The quantum average ($N_f=0$) is quite different from the image to be sent. With only $N_f\simeq 5$ known pixels picked by EO, the sleeves emerge. In contrast, the sleeves appear until 50 pixels are known if they are picked randomly. For the VO, the sleeves also emerge with $5$ pixels but in a bad shape. The shape of sleeves is reconstructed with $N_f \simeq 20$ in VO to a similar quality as $N_f \simeq 5$ in EO. The length of the sleeves is corrected with $N_f \simeq 50$ for EO and $N_f \simeq 110$ for RO and VO.

\begin{figure*}[tbp]
	\centering
	\includegraphics[angle=0,width=1\linewidth]{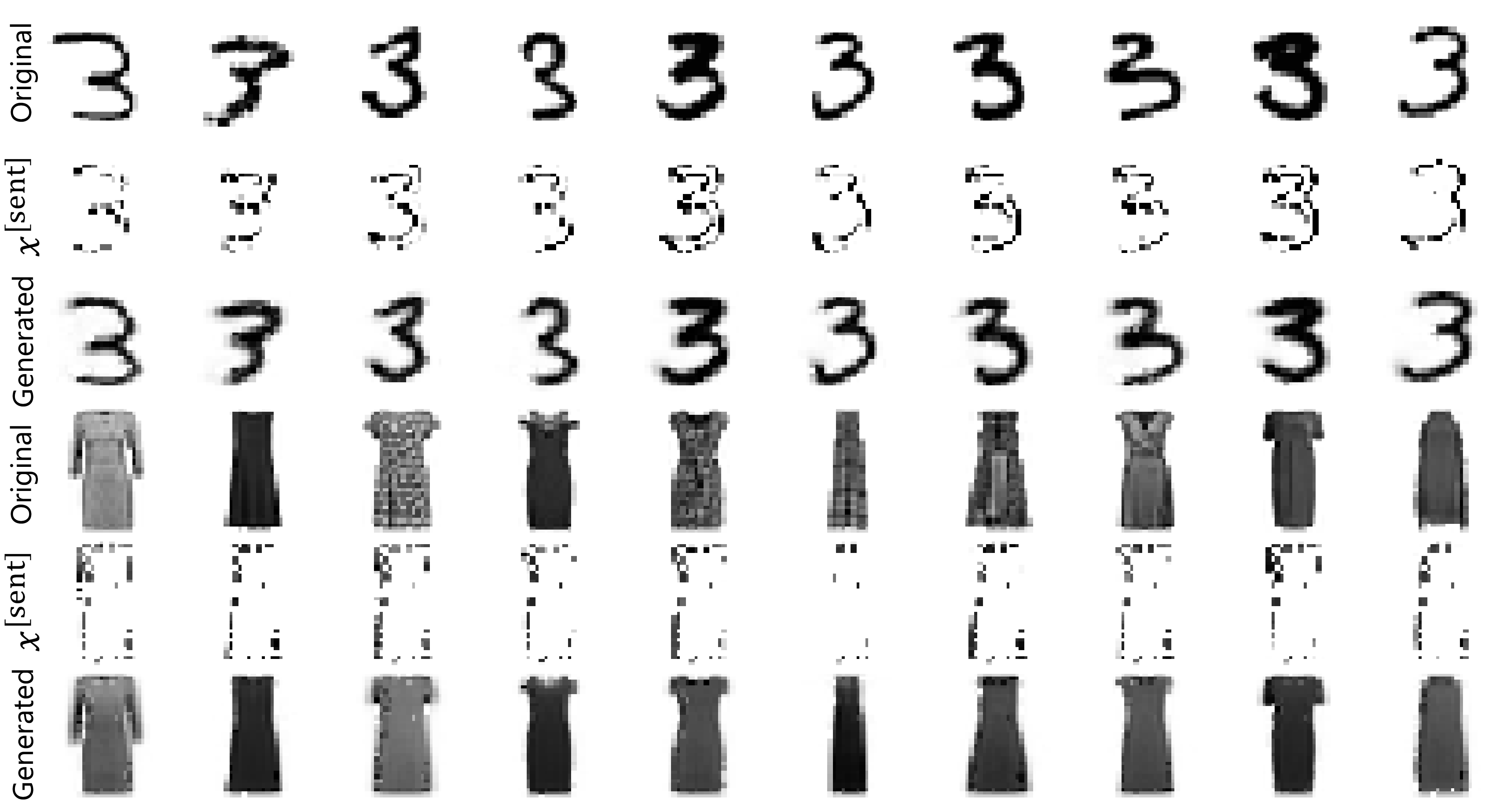}
	\caption{(Color online) Images (digits ``3'' in the first row and dresses in the forth row), the pixels $\{x^{[\text{sent}]}\}$ (the second and fifth rows), and the generated images (the third and sixth rows). We take $N_f=80$ known pixels selected by EO. The generated images in the same row are from a same state written in the form of MPS. We take the bond dimension of the MPS as $\chi=40$.}
	\label{fig-Nf80}
\end{figure*}

\begin{figure*}[tbp]
	\centering
	\includegraphics[angle=0,width=1\linewidth]{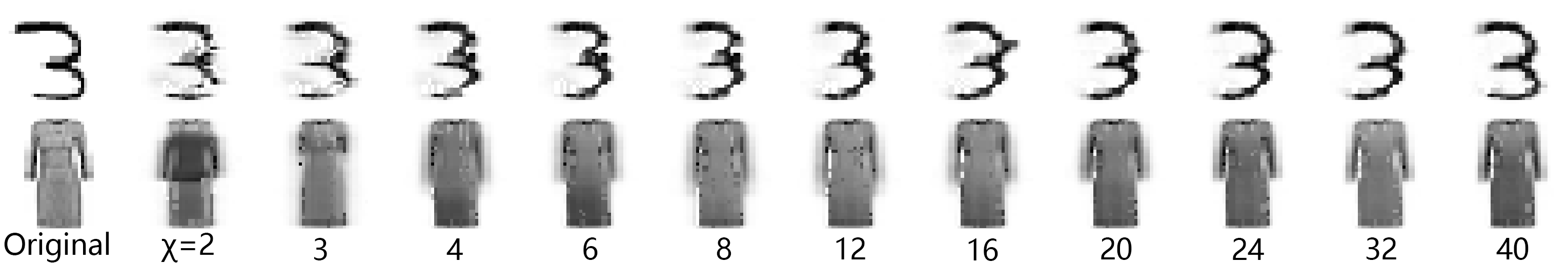}
	\caption{The original and generated images with $N_f = 80$ and different $\chi$ of the MPS.}
	\label{fig-VAryChi}
\end{figure*}

\begin{figure*}[tbp]
	\centering
	\includegraphics[angle=0,width=1\linewidth]{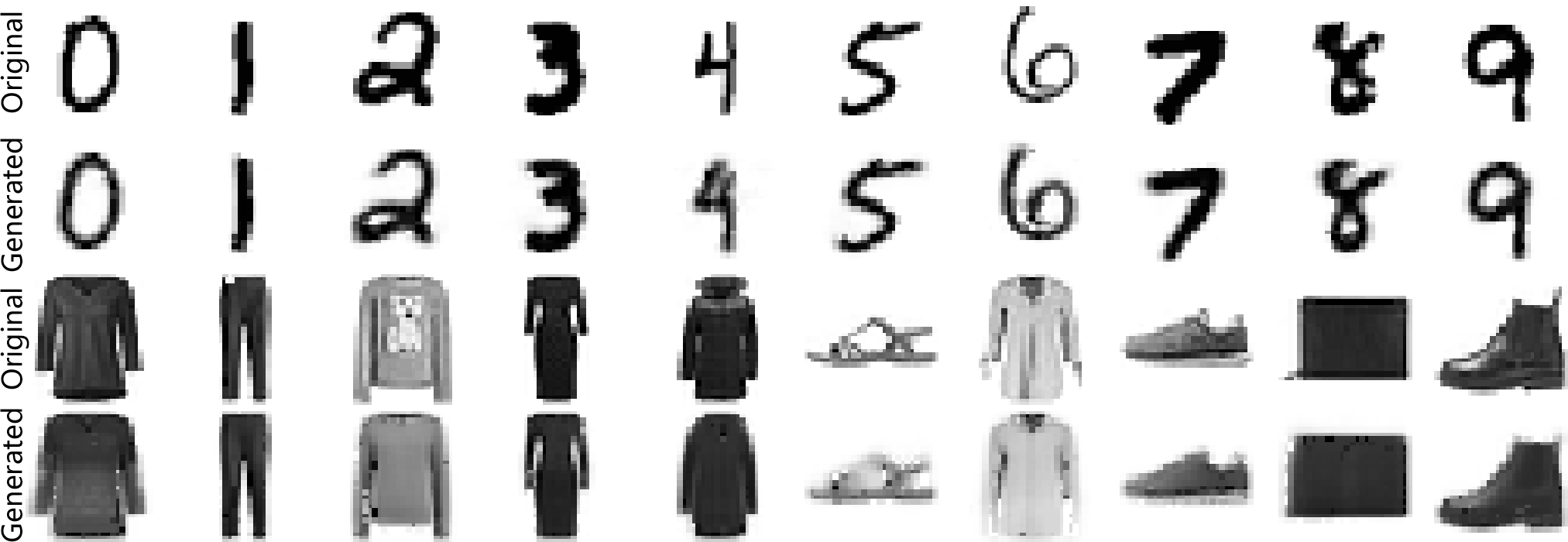}
	\caption{Original and generated images in the MNIST and fashion-MNIST datasets. We take the bond dimension of the MPS as $\chi=40$ and the number of known pixels $N_f=80$ in the EOSP.}
	\label{fig-10classes}
\end{figure*}

In Fig. \ref{fig-Nf80}, we demonstrate twenty different images from the two datasets. The first and forth rows show the original images. The second and fifth rows show $\{x^{[\text{sent}]}\}$ without being encrypted by $F$ (see the discussions in the main text). The third and sixth rows show the reconstructed images, where each image is generated from $N_f=80$ pixels selected by EOSP. Although the images (from the same dataset) are reconstructed by the same state, the differences of the shapes are well recovered. The challenging part particularly for the fashion-MNIST is to recover the details, such as the shades on the dresses.

Fig. \ref{fig-VAryChi} demonstrates the images reconstructed from the MPS's with different virtual bond dimensions $\chi$. Fig. \ref{fig-10classes} show the original and generated images from different classes of the MNIST and fashion-MNIST datasets. For each class, an MPS is trained by taking $\chi=40$. The images are generated by EOSP with $N_f=80$. In general, the quality will be improved with larger $\chi$, particularly the sharpness of the shape. However, the particular details of different images, such as the unique pictures on the coats or the stripes on the dresses, are challenging to be generated.

\bibliography{ranshiju}
\end{document}